\documentclass[runningheads]{llncs}
\usepackage[T1]{fontenc}
\usepackage[utf8]{inputenc}
\usepackage{graphicx}
\usepackage[numbers,sort&compress]{natbib}
\usepackage{algorithm}
\usepackage[noend]{algpseudocode}
\usepackage{siunitx}
\usepackage{url}
\usepackage{pifont}
\usepackage{xcolor}
\usepackage{comment}
\usepackage{multirow, makecell}
\usepackage{enumitem}
\usepackage{amsfonts}
\usepackage{booktabs}
\usepackage{needspace}
\usepackage{amsmath}
\usepackage[most]{tcolorbox}
\begin{document}
\title{Scaling Conversational Hungarian ASR: The BEA-Dialogue+ Corpus}

\titlerunning{The BEA-Dialogue+ Corpus}
\author{Máté Gedeon\inst{1,2} \and Piroska Zsófia Barta\inst{1,2} \and Péter Mihajlik\inst{1,3} \and Katalin Mády\inst{3}}
\authorrunning{M. Gedeon et al.}
\institute{Department of Telecommunications and Artificial Intelligence, \\
Budapest University of Technology and Economics, Hungary \and
Speechtex Ltd., Hungary \and
ELTE Research Centre for Linguistics, Hungary \\
\email{\{gedeonm, piroskazsofia.barta\}@edu.bme.hu, \{mihajlik.peter,mady\}@nytud.hu}}

\maketitle              

\begin{abstract}
Conversational automatic speech recognition in Hungarian is constrained by the limited amount of publicly available dialogue-style training data. The BEA-Dialogue corpus addresses this need, but its strictly speaker-disjoint train/dev/eval split reduces the usable material to only 85 hours. In this paper, we introduce BEA-Dialogue+, an expanded version of the corpus that relaxes the split criterion for experimenters and dialogue partners while preserving complete separation of the primary speakers. This results in 200 hours of transcribed natural conversations and enables a controlled study of the trade-off between additional training data and speaker overlap across the splits. We evaluate several Whisper- and FastConformer-based models on both corpus versions, including Serialized Output Training (SOT)-based fine-tuning for dialogue transcription. Our results show that the larger corpus is more challenging for models without fine-tuning, whereas SOT-based adaptation yields consistent improvements in WER, CER, cpWER, and cpCER. Overall, BEA-Dialogue+ provides a substantially larger yet still demanding benchmark for Hungarian dialogue ASR, and a practical resource for training and evaluating dialogue transcription systems.
\end{abstract}

\section{Introduction}
Hungarian automatic speech recognition (ASR)--following global trends--has made substantial progress in recent years, both in terms of the amount of available data and the sophistication of the models used \citep{hu_low_res,specom2025}. Among publicly available resources, the BEA family has played a particularly important role: the original BEA database provides large-scale spontaneous Hungarian speech, a part of which is turned into an ASR benchmark by BEA-Base and BEA-Large, together with BEA-Dialogue extending this line toward conversational settings \citep{bea2014,bea-base,bea_large}.

The BEA-Dialogue corpus was derived from the recordings of BEA, with the aim of providing speech recognition systems with natural communicative situations organized into dialogue-like units. However, a fundamental design criterion was complete speaker disjointness across the training, validation, and test sets, so that evaluation would provide a genuinely objective picture of the models' generalization ability. At the same time, this strict condition had the consequence that the effective amount of data in BEA-Dialogue was substantially reduced: the published version contains only 85 hours of conversations, compared with the original 255-hour size of BEA-Large.

This reduction may be particularly disadvantageous for modern neural-network-based speech recognition systems, whose performance depends strongly on both the amount and the diversity of the available data \citep{roger2020}. Motivated by this, in the present study we examine how much larger a dataset can be constructed by relaxing the split constraints--that is, by partially lifting full speaker disjointness--and how the performance of models trained on the resulting extended corpus compares with experiments carried out on the strictly disjoint version.

Relaxing this criterion, however, automatically raises data leakage issues. Although a specific recording can not appear in more than one split (i.e. the training, validation, and evaluation split), and we also avoided overlap of primary speakers across the splits (by primary speaker we mean the main informant, who speaks the most), experimenters and dialogue partners may still occur in multiple splits. The generalizability of the results must therefore be interpreted with caution \citep{data_leakage}. At the same time, we note that allowing speaker overlap (i.e., not filtering it out) is standard practice in settings where eliminating it is nearly impossible, such as broadcast news \citep{arsoy2007}.

The main contributions of this paper are as follows:
\begin{itemize}[leftmargin=*]
    \item We introduce BEA-Dialogue+, a 200-hour conversational Hungarian speech corpus derived from BEA recordings that preserves full separation of primary speakers while substantially increasing the amount of usable training data.
    \item We provide a detailed comparison between BEA-Dialogue and BEA-Dialogue+, including split sizes, overlap statistics, and the distribution of speaker changes, thereby quantifying how the relaxed split alters corpus difficulty.
    \item We benchmark several Whisper and FastConformer models on both corpus versions, including SOT-based fine-tuning for dialogue transcription, under a unified evaluation protocol.
    \item We analyze the trade-off between increased training data and the risk of speaker overlap across splits, showing that the enlarged corpus remains a meaningful and challenging benchmark for Hungarian dialogue transcription.
\end{itemize}

The BEA-Dialogue+ corpus used in the experiments is available for research purposes\footnote{\url{https://phon.nytud.hu/bea/}}.

\section{Related Work}
Hungarian ASR has long been shaped by the availability of relatively limited public speech resources, a situation documented in the recent quantitative survey of spoken Hungarian datasets by \citet{hu_low_res}. Within this landscape, the BEA database is one of the largest and most influential spontaneous-speech resources for Hungarian \citep{bea2014}. Subsequent work has turned parts of BEA into increasingly task-oriented benchmarks: BEA-Base focuses on spontaneous Hungarian ASR evaluation, while revised annotation guidelines have improved the consistency and usability of Hungarian speech corpora more broadly \citep{bea-base,Mady2024RevisedAnnotation}. Most recently, BEA-Dialogue shifted attention toward conversational transcription and dialogue-oriented evaluation, providing the direct starting point for the present work \citep{bea_large}.

At a broader methodological level, the ASR literature has repeatedly shown that performance is strongly influenced by both corpus size and corpus diversity \citep{roger2020}. This is especially relevant for Hungarian, where recent work has shown that increasing the amount of supervised training data can yield substantial gains even in the presence of large pre-trained models \citep{specom2025}. In terms of model families, Whisper has become a widely used large-scale baseline for robust speech recognition, while FastConformer offers an efficient architecture for strong end-to-end ASR performance \citep{whisper,fastconformer}. For overlapped and multi-speaker recognition, Serialized Output Training (SOT) provides a practical way to represent speaker changes in a single output stream, which makes it a natural choice for dialogue transcription experiments such as ours \citep{SOT}.

A separate line of related work concerns evaluation methodology and the risks associated with split design. Data leakage is now recognized as a major source of overoptimistic conclusions in machine-learning-based science, including speech and language processing workflows \citep{data_leakage}. At the same time, some realistic ASR conditions make perfect speaker separation difficult to maintain, and corpora such as broadcast news have historically tolerated speaker recurrence across data partitions when stricter constraints would render the task impractical \citep{arsoy2007}. The original BEA-Dialogue corpus deliberately prioritized strict speaker disjointness \citep{bea_large}; in contrast, the present work studies a controlled relaxation of that criterion, with the explicit goal of understanding whether the gain in usable training material outweighs the methodological risks.

\section{Data}
The BEA-Dialogue corpus was designed specifically for conversational speech processing research. It is based on recordings from 242 speakers in the BEA database \citep{bea2014} (transcribed according to \citet{Mady2024RevisedAnnotation}) who had not previously appeared in the BEA-Base database \citep{bea-base}.

During the creation of the new dataset, the utterances in the recordings were extracted together with their time stamps and speaker identifiers (SPK -- primary speaker, EXP -- experimenter, DP -- dialogue partner). The dialogues were segmented along pauses in the recordings, yielding coherent dialogue excerpts with natural boundaries. These shorter units were later merged into larger segments of approximately 30 seconds, which are well suited to automatic speech processing and language technology tasks.

In addition to the primary speakers, the corpus includes several female and male experimenters and dialogue partners, providing a variety of speaking situations and interaction patterns. In BEA-Dialogue, the training, validation (dev), and evaluation (eval) subsets are disjoint with respect to all speakers. To achieve this, the split was tied to three experimenters so that neither primary speakers nor partners were repeated across the subsets. As a consequence, for example, the \textit{discourse} module---the only module in which dialogue partners also appear---was not included in the dataset for some speakers.

The resulting BEA-Dialogue is the largest dialogue-oriented subcollection of the BEA database to date that provides a speaker-independent split. However, this independence severely limited the attainable size of the corpus.

For BEA-Dialogue+, we relaxed the constraint by requiring independence only for the primary speaker. Table~\ref{tab:dial_meta} summarizes the result. As can be seen, this increased the amount of data from 85 to 200 hours. The combined size of the \textit{dev} and \textit{eval} sets is similar in both versions, but the two are more balanced in BEA-Dialogue+.

\begin{table*}[t]
\centering
\renewcommand{\arraystretch}{1.2}
\setlength{\tabcolsep}{8pt}
\resizebox{1\columnwidth}{!}{
\begin{tabular}{l|ccc|ccc}
\toprule
& \multicolumn{3}{c}{\textbf{BEA-Dialogue}} & \multicolumn{3}{c}{\textbf{BEA-Dialogue+}} \\
\cmidrule(lr){2-4} \cmidrule(lr){5-7}
& \textbf{Train} & \textbf{Dev} & \textbf{Eval} & \textbf{Train} & \textbf{Dev} & \textbf{Eval} \\
\midrule
\textbf{\# Speakers [f|m]} & 121 | 67 & 3 | 6 & 29 | 16 & 179 | 126 & 11 | 4 & 15 | 2 \\
\textbf{\# Segments} & 9{,}179 & 577 & 1{,}906 & 25{,}193 & 1{,}084 & 1{,}207 \\
\textbf{\# Words} & 532{,}732 & 34{,}056 & 105{,}472 & 1{,}587{,}977 & 65{,}399 & 75{,}119 \\
\textbf{\# Characters} & 3{,}217{,}617 & 206{,}740 & 641{,}628 & 9{,}402{,}372 & 387{,}251 & 445{,}382 \\
\textbf{\# Nonlexical units} & 44{,}013 & 1{,}662 & 7{,}041 & 113{,}361 & 3{,}482 & 5{,}286 \\
\textbf{Avg. \# Speakers / Segment} & 1.77 & 1.92 & 1.61 & 1.97 & 1.81 & 1.94 \\
\textbf{Avg. \# Utterances / Segment} & 10.99 & 8.68 & 9.74 & 10.39 & 8.48 & 9.91 \\
\textbf{Avg. Segment Duration [s]} & 26.23 & 26.31 & 26.09 & 26.21 & 26.0 & 25.95 \\
\textbf{SPK Duration [h]} & 46.41 & 3.18 & 9.64 & 111.66 & 4.60 & 4.99 \\
\textbf{EXP Duration [h]} & 15.40 & 0.61 & 2.13 &  54.03 & 2.27 & 2.83 \\
\textbf{DP Duration [h]} & 1.39 & 0.38 & 0.76 & 16.82 & 0.76 & 0.81 \\
\textbf{Liter. Overlap Duration [h]} & 1.60 & 0.21 & 0.26 & 8.58 & 0.31 & 0.40 \\
\textbf{Total Overlap Duration [h]} & 3.28 & 0.29 & 0.40 & 14.54 & 0.44 & 0.65 \\
\midrule
\textbf{Total Duration [h]} & 66.87 & 4.22 & 13.81 & 183.41 & 7.83 & 8.70 \\
\bottomrule
\end{tabular}}
\vspace{5pt}
\caption{Corpus statistics for the two versions of BEA-Dialogue.}
\label{tab:dial_meta}
\end{table*}

\begin{figure}[ht]
\centering
\includegraphics[width=0.75\linewidth]{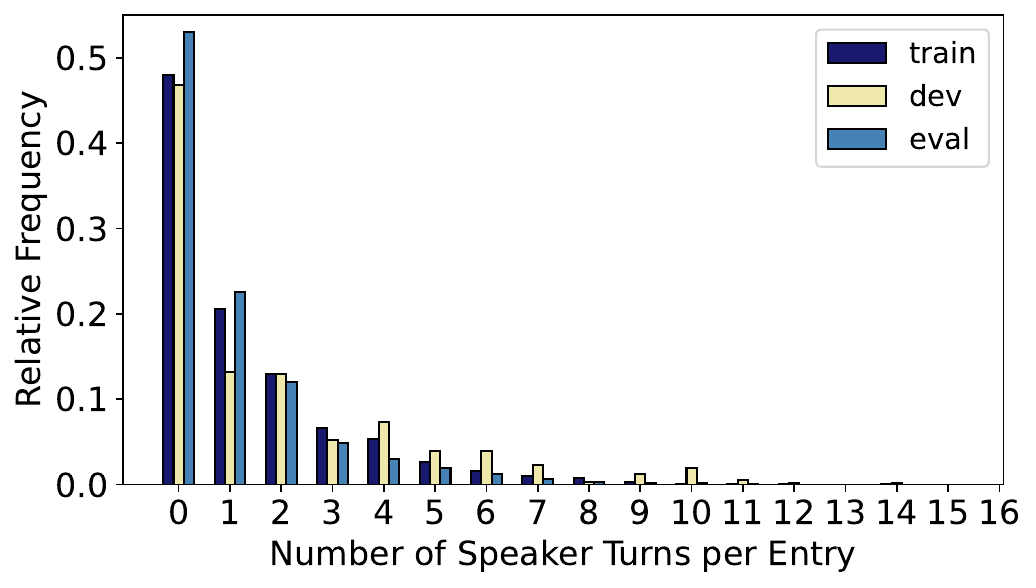}
\caption{Distribution of the number of speaker changes per segment in BEA-Dialogue.}
\label{fig:spk_changes1}
\end{figure}

\begin{figure}[ht]
\centering
\includegraphics[width=0.75\linewidth]{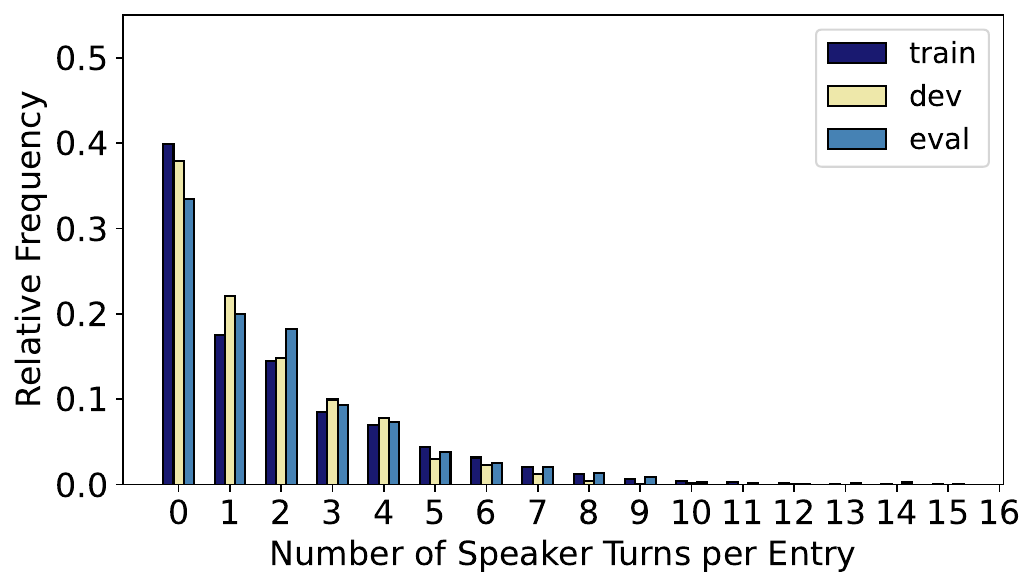}
\caption{Distribution of the number of speaker changes per segment in BEA-Dialogue+.}
\label{fig:spk_changes2}
\end{figure}

Beyond the amount of data, an interesting change is that the number of speaker changes per 30-second segment also shifts when the constraint is relaxed (Figures~\ref{fig:spk_changes1} and~\ref{fig:spk_changes2}). As can be seen, the new corpus contains fewer segments with zero or one speaker change, so the task is expected to become more difficult because overlapping speech is more likely, which particularly complicates transcription. This is confirmed by the \textit{Overlap Duration} row in Table~\ref{tab:dial_meta}, which sums all acoustic overlap generated by the speakers, and especially by the \textit{Overlap Duration} row, which sums the overlap regions that are explicitly transcribed and evaluated in the ASR experiments reported here.

\section{Experiments}
For the BEA-Dialogue+ dataset, we trained the same models as those presented in the original BEA-Dialogue paper \citep{bea_large}, and complemented them with our own in-house models. During training, we applied Serialized Output Training (SOT)~\cite{SOT}, in which speaker changes are indicated by a \texttt{<sc>} (speaker change) token. These tokens were inserted so that each speaker's utterances remain contiguous even if another speaker interrupts. In this way, the model is trained to recognize speaker boundaries while preserving the linguistic integrity of the individual utterances. For example:

\begin{quote}
\textit{Hi, did you get the letter?} \texttt{<sc>} \textit{Yes, I'm reading it right now.} \texttt{<sc>} \textit{What do you think of it?} \texttt{<sc>} \textit{It's quite interesting; I think it is worth discussing.}
\end{quote}

For training, we removed hesitations, laughter, and all other annotated non-verbal events from the database \citep{Mady2024RevisedAnnotation}. The models were trained exclusively on the cleaned verbal content and evaluated on the same type of data. In addition, for more efficient training, exceptionally long segments in the training set were excluded.

We evaluated the models using \textit{WER} (Word Error Rate) and \textit{CER} (Character Error Rate), as well as their dialogue-specific variants (cpWER and cpCER). The latter minimize errors over all possible permutations in which the units are delimited by the \texttt{<sc>} tokens. Because some segments contained more than ten speaker changes, evaluating all permutations was not feasible. To address this, we used a hybrid strategy: for up to seven speaker changes, we computed all possible permutations, whereas for larger numbers of changes we used a \textit{beam search} algorithm to obtain near-optimal results.

For the fine-tuned models, we also report speaker-change accuracy (\textit{scAcc}), which shows the percentage of cases in which the model correctly predicted the occurrences of the \texttt{<sc>} tokens. We used the NVIDIA NeMo toolkit \citep{nemo} for training.

\section{Results}
Table~\ref{tab:bea_dialogue_results} shows the results obtained on the BEA-Dialogue corpus, while Table~\ref{tab:bea_dialogue_plus_results} contains the values obtained on the extended BEA-Dialogue+ dataset. In both cases, we used identical models and a uniform evaluation protocol, making the results directly comparable. We evaluated Whisper \citep{whisper} and FastConformer \citep{fastconformer} models.

\begin{table*}[t]
\centering
\resizebox{\textwidth}{!}{
\begin{tabular}{l|ccccc|cccccc}
\toprule
\multirow{2}{*}{\textbf{Model}} & \multicolumn{5}{c}{\textbf{dev}} & & \multicolumn{5}{c}{\textbf{eval}} \\
\cmidrule(lr){2-6} \cmidrule(lr){8-12}
 & \textbf{WER} & \textbf{cpWER} & \textbf{CER} & \textbf{cpCER} & \textbf{scAcc} & &
   \textbf{WER} & \textbf{cpWER} & \textbf{CER} & \textbf{cpCER} & \textbf{scAcc} \\
\midrule
\texttt{whisper-medium}     & 25.45 & 25.27 & 12.61 & 12.42 & -- & & 29.21 & 29.12 & 14.71 & 14.63 & -- \\
\texttt{whisper-large-v2}   & 19.65 & 19.42 & 9.84  & 9.58  & -- & & 24.50 & 24.42 & 13.13 & 13.05 & -- \\
\texttt{whisper-large-v3}   & 21.19 & 21.04 & 12.74 & 12.56 & -- & & 22.21 & 22.13 & 12.27 & 12.18 & -- \\
\texttt{fc\_hu\_l (zs)}    & 14.75 & 14.60 & 6.37 & 6.23 & -- & & 16.33 & 16.27 & 7.76 & 7.71 & -- \\
\texttt{fc\_hu\_xl (zs)}   & 13.27 & 13.17 & 5.89 & 5.76 & -- & & 15.48 & 15.43 & 7.47 & 7.42 & -- \\
\midrule
\texttt{fc\_en\_l (ft)}     & 19.69 & 19.53 & 7.95 & 7.78 & 69.32 & & 20.56 & 20.44 & 9.11 & 9.00 & \textbf{82.16} \\
\texttt{fc\_hu\_l (ft)}     & 12.19 & 11.96 & 5.63 & 5.45 & 67.42 & & 13.90 & 13.80 & 7.03 & 6.93 & 79.20 \\
\texttt{fc\_hu\_xl (ft)}    & \textbf{11.43} & \textbf{11.21} & \textbf{5.32} & \textbf{5.12} & \textbf{70.71} & &
                              \textbf{13.03} & \textbf{12.92} & \textbf{6.65} & \textbf{6.55} & 80.94 \\
\bottomrule
\end{tabular}}
\vspace{5pt}
\caption{ASR results on BEA-Dialogue (\%).}
\label{tab:bea_dialogue_results}
\end{table*}

Table~\ref{tab:bea_dialogue_results} contains the baseline values reported by \citet{bea_large} (with the fine-tuned model denoted as \texttt{fc\_en\_l (ft)}), which we supplemented with two additional FastConformer CTC Large and XLarge models pre-trained on Hungarian data following the recipe of \citet{specom2025}. One of these models (\texttt{fc\_hu\_l}) is architecturally identical to the FastConformer CTC Large model fine-tuned from English weights, whereas the other (\texttt{fc\_hu\_xl}) is a larger FastConformer CTC XL model. These models were pre-trained on 2,700 hours of Hungarian speech, and as a result they achieved substantially better performance than the Whisper models even without fine-tuning; they even produced more favorable results than the FastConformer CTC model fine-tuned from English weights. In the tables, fine-tuned models are marked with \texttt{(ft)}, while those without fine-tuning (zero-shot) are marked with \texttt{(zs)}. The post-fine-tuning results further reinforce the relevance of the corpus: even models originally trained on Hungarian benefited from dialogue-specific continued training, highlighting the value of the BEA-Dialogue corpus for the development of dialogue transcription models.

\begin{table*}[t]
\centering
\resizebox{\textwidth}{!}{
\begin{tabular}{l|ccccc|cccccc}
\toprule
\multirow{2}{*}{\textbf{Model}} & \multicolumn{5}{c}{\textbf{BEA-Dialogue+-dev}} & & \multicolumn{5}{c}{\textbf{BEA-Dialogue+-eval}} \\
\cmidrule(lr){2-6} \cmidrule(lr){8-12}
 & \textbf{WER} & \textbf{cpWER} & \textbf{CER} & \textbf{cpCER} & \textbf{scAcc} & &
   \textbf{WER} & \textbf{cpWER} & \textbf{CER} & \textbf{cpCER} & \textbf{scAcc} \\
\midrule
\texttt{whisper-medium}     & 30.83 & 30.73 & 15.96 & 15.83 & -- & & 30.19 & 30.06 & 16.00 & 15.86 & -- \\
\texttt{whisper-large-v2}   & 24.82 & 24.70 & 13.46 & 13.30 & -- & & 25.48 & 25.36 & 14.68 & 14.52 & -- \\
\texttt{whisper-large-v3}   & 23.28 & 23.15 & 13.15 & 12.99 & -- & & 23.27 & 23.17 & 13.34 & 13.22 & -- \\
\texttt{fc\_hu\_l (zs)}    & 18.07 & 17.99 & 8.11  & 8.02  & -- & & 18.91 & 18.80 & 9.50  & 9.40  & -- \\
\texttt{fc\_hu\_xl (zs)}   & 16.76 & 16.68 & 7.68  & 7.59  & -- & & 17.32 & 17.19 & 9.00  & 8.88  & -- \\
\midrule
\texttt{fc\_en\_l (ft)}     & 16.30 & 16.11 & 7.42  & 7.23  & \textbf{73.05} & & 16.49 & 16.28 & 8.29  & 8.11  & \textbf{69.11} \\
\texttt{fc\_hu\_l (ft)}     & 14.97 & 14.78 & 7.12  & 6.94  & 70.64 & & 15.19 & 14.99 & 7.95  & 7.79  & 67.99 \\
\texttt{fc\_hu\_xl (ft)}    & \textbf{12.84} & \textbf{12.67} & \textbf{6.31} & \textbf{6.15} & \textbf{73.05} & &
                              \textbf{13.59} & \textbf{13.42} & \textbf{7.34} & \textbf{7.18} & 67.56 \\
\bottomrule
\end{tabular}}
\vspace{5pt}
\caption{ASR results on BEA-Dialogue+ (\%).}
\label{tab:bea_dialogue_plus_results}
\end{table*}

Based on the BEA-Dialogue+ results (see Table~\ref{tab:bea_dialogue_plus_results}), it is clear that models without fine-tuning perform worse in every case than on BEA-Dialogue, typically showing at least a 10\% relative degradation. One explanation is the speaker-change distribution shown in Figure~\ref{fig:spk_changes2}: whereas BEA-Dialogue contains many segments in which zero or only one change occurs, BEA-Dialogue+ contains more complex segments with multiple changes, especially in the \textit{eval} set. This increases task difficulty for the model, particularly when the model is not given explicit information about speaker-change boundaries.

By contrast, fine-tuning leads to larger improvements on this corpus, for several possible reasons. First, because of the much larger amount of training data, fine-tuned models can adapt more effectively to the characteristics of BEA-Dialogue+, and thus handle more complex segments with more speaker changes more successfully. Second, it cannot be ruled out that speaker overlap introduced during the expansion of BEA-Dialogue (i.e., data leakage) also contributed to this improvement, because it may make it easier for the model to recognize the voice of a given speaker. After fine-tuning, the pre-trained models achieved particularly large relative gains on this corpus, but they still fell short of the results measured on BEA-Dialogue. Thus, despite the larger amount of data, the task did not become easier, and the corpus remains similarly useful for system comparison.

For both corpora, cpWER and cpCER are consistently lower than WER and CER. The magnitude of the improvement varies, but it is generally larger for the fine-tuned models.

The behavior of the scAcc metric is related to the error rates: in more difficult segments (with more speaker changes), the accuracy of speaker-change recognition deteriorates in parallel with the increase in WER. This suggests that scAcc reflects not only the success of speaker-boundary detection, but also indirectly indicates task difficulty. At the same time, a better scAcc does not necessarily yield a better WER, as shown by the \texttt{fc\_en\_l (ft)} and \texttt{fc\_hu\_l (ft)} models: although the former is better at identifying speaker changes, the latter is better at transcription.

In BEA-Dialogue+, the speaker overlap introduced during corpus expansion may have a favorable effect on the performance of the fine-tuned models. If the model has already encountered earlier occurrences of the same speaker in the training data, it can more easily adapt to that voice and its acoustic characteristics. Quantitative investigation of this effect requires further analysis; however, the results suggest that the improvement can be explained only partly by data leakage, since the largest gains appear in complex segments, where modeling dialogue structure is more likely to be the decisive factor.

Table~\ref{tab:compare} illustrates a concrete example from the BEA-Dialogue+ \textit{eval} set with the outputs of the different models. The audio file is four seconds long and comes from the end of a conversation. Despite its short duration, many words are spoken, most of them overlapping, and only the fine-tuned models are visibly able to handle this case.

\begin{tcolorbox}[
    title={Table 4: Comparison on a concrete example from the BEA-Dialogue+ eval set},
    colback=white,
    colframe=black!60,
    float*=ht,
    halign=left,
    sharp corners
]\label{tab:compare}
\small
\textbf{Reference:}\\[2pt]
\textit{hát igen jó köszönöm szépen szerintem elég lesz ennyi <sc> köszönjük <sc> kösz} \\
\vspace{6pt}
\renewcommand{\arraystretch}{1.35}
\begin{tabular}{p{2.8cm} p{5.6cm} r r}
\textbf{Model} & \textbf{Transcript} & \textbf{WER} & \textbf{CER} \\
\hline
\texttt{whisper-medium}   & \textit{jó köszönöm szépen szépen szerep} & 72.73 & 60.29 \\  & \textit{} & & \\
\texttt{whisper-large-v2}   & \textit{igen jó köszönöm szépen köszönjük} & 54.55 & 51.47 \\  & \textit{} & & \\
\texttt{whisper-large-v3}   & \textit{hát igen jó okosan szépen szívesen köszönjük} & 54.55 & 44.12 \\  & \textit{} & & \\
\texttt{fc\_hu\_l (zs)}   & \textit{hát igen jó köszönöm szépen szépen} & 54.55 & 51.47 \\  & \textit{} & & \\
\texttt{fc\_hu\_xl (zs)}   & \textit{hát igen jó köszönöm szépen} & 54.55 & 60.29 \\  & \textit{} & & \\
\texttt{fc\_en\_l (ft)}   & \textit{hát igen jó köszönöm szépen szerintem elég lesz} & 27.27 & 30.88 \\  & \textit{} & & \\
\texttt{fc\_hu\_l (ft)}  & \textit{hát igen jó köszönöm szépen szerintem elég lesz ennyi <sc> kösz} & \textbf{9.09} & \textbf{14.71} \\  & \textit{} & & \\
\texttt{fc\_hu\_xl (ft)}  & \textit{hát igen jó köszönöm szépen szerintem elég lesz ennyi <sc> kösz} & \textbf{9.09} & \textbf{14.71} \\ & \textit{} & & \\
\end{tabular}
\end{tcolorbox}

\section{Conclusion}
In this study, we introduced the BEA-Dialogue+ corpus, which represents a substantial quantitative expansion over the earlier BEA-Dialogue dataset for Hungarian conversational ASR research. The primary motivation behind constructing the corpus was to significantly increase the amount of available data by partially relaxing the strict speaker-disjointness constraints while preserving complete separation of the primary speakers. The result is a 200-hour corpus, expanding the earlier 85-hour version by nearly a factor of 2.5.

The experimental results show that corpus expansion has a dual effect. The performance of pre-trained models without fine-tuning degrades by about 10\% on BEA-Dialogue+ compared with BEA-Dialogue, largely due to the increased number of speaker changes. In contrast, FastConformer models fine-tuned with the Serialized Output Training (SOT) method show larger relative improvements than on BEA-Dialogue.

The main advantage of the corpus is the substantially increased amount of data, which enables more effective training of dialogue transcription systems. In addition, the corpus segmented into uniform units (targeting segments of about 30 seconds) is well suited to system evaluation and comparison. The precise extent of the potential data leakage caused by speaker overlap requires further investigation, for example through evaluation on a set independent of those used so far, which would provide a more accurate picture of both the effect of increasing data volume and the impact of speaker overlap. Both BEA-Dialogue and BEA-Dialogue+ are expected to become downloadable for registered researchers in the near future and can thus be used for Conversational AI research.

\section*{Acknowledgments}
\label{Koszonet}
This work was supported by the Ministry of Culture and Innovation of Hungary from the National Research, Development and Innovation Fund under the EKÖP\_KDP-25-1-BME-21 funding scheme, by Project No.~2025-2.1.2-EKÖP-KDP-2025-00005, and by NKFIH projects K143075 and K135038.


%
%
%
\bibliographystyle{plainnat}
\bibliography{main}

@inproceedings{SOT,
  title={Serialized Output Training for End-to-End Overlapped Speech Recognition},
  author={Naoyuki Kanda and Yashesh Gaur and Xiaofei Wang and Zhong Meng and Takuya Yoshioka},
  booktitle={Interspeech},
  year={2020},
  url={https://api.semanticscholar.org/CorpusID:214714409}
}

@inproceedings{bea-base,
    title = "{BEA}-Base: A Benchmark for {ASR} of Spontaneous {H}ungarian",
    author = "Mihajlik, Peter  and
      Balog, Andras  and
      Graczi, Tekla Etelka  and
      Kohari, Anna  and
      Tarj{\'a}n, Bal{\'a}zs  and
      Mady, Katalin",
    editor = "Calzolari, Nicoletta  and
      B{\'e}chet, Fr{\'e}d{\'e}ric  and
      Blache, Philippe  and
      Choukri, Khalid  and
      Cieri, Christopher  and
      Declerck, Thierry  and
      Goggi, Sara  and
      Isahara, Hitoshi  and
      Maegaard, Bente  and
      Mariani, Joseph  and
      Mazo, H{\'e}l{\`e}ne  and
      Odijk, Jan  and
      Piperidis, Stelios",
    booktitle = "Proceedings of the Thirteenth Language Resources and Evaluation Conference",
    month = jun,
    year = "2022",
    address = "Marseille, France",
    publisher = "European Language Resources Association",
    url = "https://aclanthology.org/2022.lrec-1.211/",
    pages = "1970--1977"
}

@article{data_leakage,
title = {Leakage and the reproducibility crisis in machine-learning-based science},
journal = {Patterns},
volume = {4},
number = {9},
pages = {100804},
year = {2023},
issn = {2666-3899},
doi = {https://doi.org/10.1016/j.patter.2023.100804},
url = {https://www.sciencedirect.com/science/article/pii/S2666389923001599},
author = {Sayash Kapoor and Arvind Narayanan}
}

@misc{bea_large,
      title={Toward Conversational Hungarian Speech Recognition: Introducing the {BEA-L}arge and {BEA-D}ialogue Datasets}, 
      author={Máté Gedeon and Piroska Zsófia Barta and Péter Mihajlik and Tekla Etelka Gráczi and Anna Kohári and Katalin Mády},
      year={2025},
      eprint={2511.13529},
      archivePrefix={arXiv},
      primaryClass={cs.CL},
      url={https://arxiv.org/abs/2511.13529}, 
}

@InProceedings{specom2025,
author="Dobsinszki, Gergely
and Mihajlik, P{\'e}ter
and K{\'a}d{\'a}r, M{\'a}t{\'e} Soma
and Fegy{\'o}, Tibor
and M{\'a}dy, Katalin",
editor="Karpov, Alexey
and Gosztolya, G{\'a}bor",
title="Best Data is more Supervised Data -- Even for Hungarian ASR",
booktitle="Speech and Computer",
year="2026",
publisher="Springer Nature Switzerland",
address="Cham",
pages="60--69",
isbn="978-3-032-07959-6"
}

@inproceedings{hu_low_res,
    title = "Is Spoken {H}ungarian Low-resource?: A Quantitative Survey of {H}ungarian Speech Data Sets",
    author = {Mihajlik, Peter  and
      M{\'a}dy, Katalin  and
      Koh{\'a}ri, Anna  and
      Fruzsina, Fruzsina S{\'a}ra  and
      Kiss, G{\'a}bor  and
      Gr{\'a}czi, Tekla Etelka  and
      Do{\u{g}}ru{\"o}z, A. Seza},
    editor = "Calzolari, Nicoletta  and
      Kan, Min-Yen  and
      Hoste, Veronique  and
      Lenci, Alessandro  and
      Sakti, Sakriani  and
      Xue, Nianwen",
    booktitle = "Proceedings of the 2024 Joint International Conference on Computational Linguistics, Language Resources and Evaluation (LREC-COLING 2024)",
    month = may,
    year = "2024",
    address = "Torino, Italia",
    publisher = "ELRA and ICCL",
    url = "https://aclanthology.org/2024.lrec-main.820/",
    pages = "9382--9388"
}

@misc{roger2020,
      title={Deep Neural Networks for Automatic Speech Processing: A Survey from Large Corpora to Limited Data}, 
      author={Vincent Roger and Jérôme Farinas and Julien Pinquier},
      year={2020},
      eprint={2003.04241},
      archivePrefix={arXiv},
      primaryClass={eess.AS},
      url={https://arxiv.org/abs/2003.04241}, 
}

@inproceedings{arsoy2007,
  title     = {Language modeling for automatic turkish broadcast news transcription},
  author    = {Ebru Arısoy and Haşim Sak and Murat Saraçlar},
  year      = {2007},
  booktitle = {Interspeech 2007},
  pages     = {2381--2384},
  doi       = {10.21437/Interspeech.2007-273},
  issn      = {2958-1796},
}

@misc{nemo,
      title={NeMo: a toolkit for building AI applications using Neural Modules}, 
      author={Oleksii Kuchaiev and Jason Li and Huyen Nguyen and Oleksii Hrinchuk and Ryan Leary and Boris Ginsburg and Samuel Kriman and Stanislav Beliaev and Vitaly Lavrukhin and Jack Cook and Patrice Castonguay and Mariya Popova and Jocelyn Huang and Jonathan M. Cohen},
      year={2019},
      eprint={1909.09577},
      archivePrefix={arXiv},
      primaryClass={cs.LG},
      url={https://arxiv.org/abs/1909.09577}, 
}

@misc{fastconformer,
      title={Fast Conformer with Linearly Scalable Attention for Efficient Speech Recognition}, 
      author={Dima Rekesh and Nithin Rao Koluguri and Samuel Kriman and Somshubra Majumdar and Vahid Noroozi and He Huang and Oleksii Hrinchuk and Krishna Puvvada and Ankur Kumar and Jagadeesh Balam and Boris Ginsburg},
      year={2023},
      eprint={2305.05084},
      archivePrefix={arXiv},
      primaryClass={eess.AS},
      url={https://arxiv.org/abs/2305.05084}, 
}

@misc{whisper,
      title={Robust Speech Recognition via Large-Scale Weak Supervision}, 
      author={Alec Radford and Jong Wook Kim and Tao Xu and Greg Brockman and Christine McLeavey and Ilya Sutskever},
      year={2022},
      eprint={2212.04356},
      archivePrefix={arXiv},
      primaryClass={eess.AS},
      url={https://arxiv.org/abs/2212.04356}, 
}

@InProceedings{bea2014,
author="Neuberger, Tilda
and Gyarmathy, Dorottya
and Gr{\'a}czi, Tekla Etelka
and Horv{\'a}th, Vikt{\'o}ria
and G{\'o}sy, M{\'a}ria
and Beke, Andr{\'a}s",
editor="Sojka, Petr
and Hor{\'a}k, Ale{\v{s}}
and Kope{\v{c}}ek, Ivan
and Pala, Karel",
title="Development of a Large Spontaneous Speech Database of Agglutinative Hungarian Language",
booktitle="Text, Speech and Dialogue",
year="2014",
publisher="Springer International Publishing",
address="Cham",
pages="424--431",
isbn="978-3-319-10816-2"
}

@article{Mady2024RevisedAnnotation,
  author    = {M{\'a}dy, Katalin and Gr{\'a}czi Tekla Etelka and Koh{\'a}ri, Anna and Mihajlik, P{\'e}ter},
  title     = {Revised annotation conventions in Hungarian speech corpora},
  journal   = {Besz{\'e}dtudom{\'a}ny / Speech Science},
  volume    = {4},
  number    = {1},
  pages     = {185--202},
  year      = {2024},
  note      = {18 p.},
  doi       = {}
}
\end{document}